\documentclass[preprints,article,accept,moreauthors,pdftex]{Definitions/mdpi}

\firstpage{1} 
\pubvolume{1}
\issuenum{1}
\articlenumber{0}
\pubyear{2024}
\copyrightyear{2024}
\datereceived{ } 
\daterevised{ } 
\dateaccepted{ } 
\datepublished{ } 
\hreflink{https://doi.org/} 
\pdfoutput=1 

\Title{Transforming Traditional Neural Networks into Neuromorphic Quantum-Cognitive Models: A Tutorial with Applications}

\TitleCitation{Transforming Traditional Neural Networks into Neuromorphic Quantum-Cognitive Models: A Tutorial with Applications}

\definecolor{my_green}{rgb}{0.55, 0.71, 0.0}

\newcommand{\transpose}{^{\top}}
\newcommand{\imagi}{\mathsf{i}}

\usepackage{braket}
\usepackage{comment}
\usepackage{gensymb}

\Author{Milan Maksimovic\orcidA{} and Ivan S.~Maksymov\orcidB{}$^{*}$}

\AuthorNames{Milan Maksimovic and Ivan S.~Maksymov}

\AuthorCitation{Maksimovic, M.; Maksymov, I.~S.}

\address{%
Artificial Intelligence and Cyber Futures Institute, Charles Sturt University, Bathurst, NSW 2795, Australia.}

\corres{Correspondence: mmaksimovic@csu.edu.au, imaksymov@csu.edu.au}

\abstract{Quantum technologies are increasingly pervasive, underpinning the operation of numerous electronic, optical and medical devices. Today, we are also witnessing rapid advancements in quantum computing and communication. However, access to quantum technologies in computation remains largely limited to professionals in research organisations and high-tech industries. This paper demonstrates how traditional neural networks can be transformed into neuromorphic quantum models, enabling anyone with a basic understanding of undergraduate-level machine learning to create quantum-inspired models that mimic the functioning of the human brain---all using a standard laptop. We present several examples of these quantum machine learning transformations and explore their potential applications, aiming to make quantum technology more accessible and practical for broader use. The examples discussed in this paper include quantum-inspired analogues of feedforward neural networks, recurrent neural networks, Echo State Network reservoir computing and Bayesian neural networks, demonstrating that a quantum approach can both optimise the training process and equip the models with certain human-like cognitive characteristics.}


\begin{document}

\section{Introduction}
In his address to the United Nations Security Council, Professor Yann LeCun, Chief AI Scientist at Meta, highlighted significant limitations of current artificial intelligence (AI) systems \cite{LeCun}. He noted that state-of-the-art AI systems lack a true understanding of the real world, possess no persistent memory and are incapable of reasoning or effective planning. Furthermore, they fall short in acquiring new skills with the speed and efficiency demonstrated by humans or even animals.

LeCun also emphasised that intelligence cannot be achieved merely by combining extensive datasets with high-performance computing capabilities. Instead, he argued that genuine AI must arise from cognition and human-like behavioural capabilities, surpassing the superficial ability to process vast amounts of data at high speed.

While LeCun's vision, along with that of other scientists, champions the development of the next generation of cognitive and more human-like AI, the general public and many professionals not directly engaged in AI research or the application of AI technologies approach this idea with a certain degree of scepticism \cite{Gun22, Ger23, Wal25}. Notably, there is a widespread perception that the increased reliance on advanced AI tools may contribute to a decline in critical thinking and even broader cognitive skills among humans \cite{Ger25}. Moreover, the rapid proliferation of AI across various spheres of social and financial affairs presents significant challenges to the modern legal and financial systems \cite{Sou18, Mot18, Gov22}.

In this context, quantum cognition theory (QCT) models provide a novel framework for developing human-like cognitive AI by offering insights into the probabilistic and contextual nature of human thought and decision-making \cite{Khr06, Atm10, Bus12, Pot22}. Unlike classical approaches, which are often grounded in deterministic models, QCT integrates principles from quantum mechanics with psychology \cite{Bus12, Pot22}, behaviour science and decision-making, shedding light on how humans process information, perceive ambiguities and make judgments under uncertainty \cite{Mak24_information, Mak24_information1, Mak24_illusions, Mak24_APL}.

One key feature of QCT is the concept of quantum superposition that enables individuals to hold multiple, sometimes contradictory, beliefs or percepts simultaneously until a decision collapses these possibilities into a single outcome \cite{Khr06, Atm10, Bus12, Pot22}. This capability mirrors the complex and nuanced ways humans approach decision-making, particularly in scenarios involving uncertainty or conflicting information \cite{Mak24_information, Mak24_information1}. Moreover, QCT provides mechanisms for understanding phenomena such as biases, optical illusions and the contextuality of human judgments \cite{Bus12, Pot22, Mak24_illusions, Mak24_APL}, which are critical for designing AI systems that can better interpret and respond to human behaviour. 

Thus, in essence, QCT bridges the gap between cognitive psychology and advanced AI, offering a paradigm shift in how we design intelligent systems. Such systems hold the promise to enhance decision-making in real-world applications, ranging from healthcare and finance to autonomous systems \cite{Abb24_1}, by fostering a deeper integration of human-like cognitive processes within AI systems. 

It is worth noting that QCT and its practical applications have often faced unwarranted scepticism and criticism, largely due to confusion with the theory of the quantum mind \cite{Penrose}, which proposes that quantum effects may influence cognitive processes \cite{Geo_book} (see, e.g., the relevant discussion in Ref.~\cite{Bro24}). Nevertheless, the quantum mind hypothesis merits at least theoretical consideration, as it expands general knowledge and potentially contributes to a deeper understanding of the concept of quantum neural networks.

\subsection{Quantum Neural Networks}
Quantum neural networks (QNNs) are computational models that integrate the principles of quantum mechanics with neural network architectures. The concept of quantum neural computation can be traced back to the 1990s \cite{Chr95, Kak95}, though its origins may be even earlier (for a relevant review see, e.g.,~Ref.~\cite{Sch14}). Interestingly, a connection can also be drawn between the early concepts of QNNs and the theory of the quantum mind \cite{Penrose}, as well as proposals to investigate quantum effects and their potential impact on cognitive processes \cite{Geo_book}.

Contemporary research on QNNs focuses on integrating classical artificial neural networks, which are widely used in machine learning (ML) and computer vision, with the unique advantages of quantum information processing \cite{Hor09, Chi19}. This integration aims to develop more efficient algorithms by using quantum phenomena such as superposition and entanglement to enhance computational performance \cite{Ezh00, Sch14}. For instance, the work Ref.~\cite{Wan17} examines the quantum generalisation of the traditional neural network models, while Refs.~\cite{Bee20, Zha21, Hie24} introduce diverse approaches to quantum-inspired deep learning \cite{Yan21, Pir24}. Further systematic review of QNN models can be found in Refs.~\cite{Per24_2, Cho24}, including hardware implementations of QNNs.

Moreover, QNN models provide robust frameworks for developing uncertainty-aware neural networks, thereby enhancing the reliability and safety of AI systems in high-stakes applications. Yet, while quantum mind theory and related scientific hypotheses have often been regarded as distinct from the framework of QCT, as discussed, for example, in Ref.~\cite{Mak24_APL}, these approaches should eventually converge, further advancing the ability of QNNs to successfully complete these ambitious tasks.

\subsection{Neuromorphic Computing}
QNNs also share a conceptual background with neuromorphic computing (NC)  through their foundational aim of understanding and artificially reproducing cognitive processes observed in a biological brain \cite{Luk09, Tan19, Nak20, Gau21}. NC systems seek to develop hardware and algorithms inspired by the architecture and functioning of biological neural systems, thereby mimicking spiking neurons and synaptic connections to emulate the efficiency and adaptability of a biological brain \cite{Tan19, Mak23_review, Abb24_1, Ken24}. Both QNNs and NC systems aim to overcome the limitations of conventional computing architectures, which rely on deterministic logic and sequential processing and, therefore, can be suboptimal for complex tasks such as pattern recognition, real-time decision-making and adaptive learning. In particular, NC systems address these challenges by using event-driven architectures and parallel information processing, aligning with how neurons and synapses function in the brain \cite{Luk09, Tan19, Nak20}.

QNNs further enhance the concept of NC by exploiting quantum states to encode and process information, enabling parallelism and exponential scalability in specific computational tasks. Therefore, the integration of QNNs with neuromorphic principles offers exciting possibilities. For instance, NC systems could serve as a physical platform for implementing QNNs, using analogue circuits to represent quantum-inspired operations like superposition and entanglement. Conversely, QNNs could inspire new neuromorphic designs, employing quantum algorithms to model phenomena such as associative memory and probabilistic reasoning.
\begin{figure}[t]
\centering
\includegraphics[width=1.0\columnwidth]{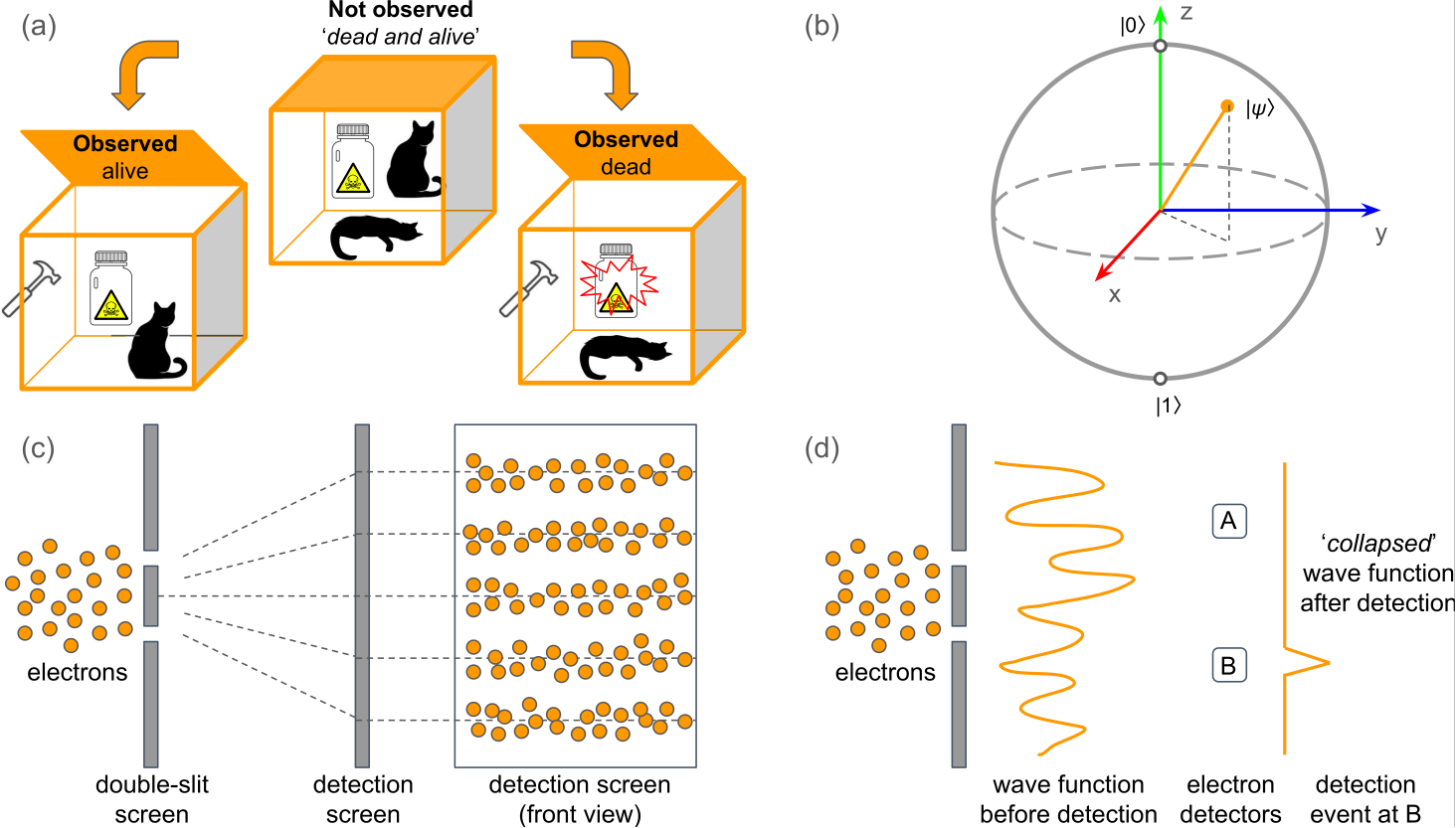}
\caption{{\bf(a)}~Schr{\"o}dinger’s cat thought experiment. A cat is placed in a sealed, opaque box containing a radioactive atom, a Geiger counter, a vial of poison and a hammer. If the atom decays, the Geiger counter triggers the hammer to release the poison, killing the cat. Until the box is opened and observed, quantum mechanics suggests the cat exists in a superposition of being both alive and dead. {\bf(b)}~Illustration of a projective measurement of a qubit \( |\psi\rangle \) using the Bloch sphere, where measurement collapses the qubit from a superposition to a definite state. {\bf(c)}~The double-slit experiment, which demonstrates quantum interference, showing that particles such as electrons behave as waves, creating an interference pattern until observed. {\bf(d)}~Illustration of wavefunction collapse triggered by detection using an electron detector.}
\label{Fig1}
\end{figure}

\subsection{Objectives of This Tutorial}
Thus, this tutorial paper aims to bridge the gap between quantum technology and broader accessibility by demonstrating how traditional neural networks can be transformed into neuromorphic quantum models. While quantum technologies are becoming increasingly pervasive in electronic, optical and medical devices---and quantum computing and communication are advancing rapidly---access to them remains largely confined to researchers and professionals in high-tech industries. This paper illustrates how anyone with a basic understanding of undergraduate-level ML can develop quantum-inspired models that emulate human brain function, all using a standard laptop.

We present several examples of quantum transformations and explore their potential applications, making quantum technology more accessible and practical. Specifically, we introduce quantum-inspired analogues of feedforward neural networks (FNNs), recurrent neural networks (RNNs), Echo State Network (ESN) reservoir computing and Bayesian neural networks (BNNs). These quantum approaches not only optimise the training process but also imbue the models with certain human-like cognitive characteristics, opening new avenues for quantum-enhanced artificial intelligence.

\section{Motivation}
\subsection{General Background}
Mechanics, a fundamental branch of classical physics, investigates the motion of physical objects resulting from the application of forces \cite{Lan76}. These objects range from macroscopic entities, such as balls and vehicles, to celestial bodies. In contrast, quantum mechanics provides a framework for describing the behaviour of systems at atomic and subatomic scales \cite{Mes62, Gri04}. Quantum mechanics extends beyond the explanatory power of classical physics, enabling the understanding of phenomena involving photons, electrons and other quantum particles \cite{Mes62, Gri04}. Furthermore, it underpins a wide array of technologies integral to contemporary society, including semiconductor devices, medical imaging systems, optical fibre communication networks and quantum computing \cite{Nie02}.

The principles of quantum mechanics often challenge intuition due to their departure from the classical framework \cite{Fey_quantum}. A prominent example is the concept of superposition that allows a quantum system to exist simultaneously in multiple states until measured \cite{Mes62, Fey_quantum, Gri04}. This concept is famously illustrated by Schr{\"o}dinger’s cat thought experiment (Figure~\ref{Fig1}a). Interestingly, Schr{\"o}dinger did not intend the notion of a dead-and-alive cat to be taken as a serious possibility. Rather, he used this thought experiment to highlight the absurdity of the prevailing interpretation of quantum mechanics \cite{Sch35}. However, advancements in quantum mechanics since Schr{\"o}dinger’s time have led scientists to propose alternative interpretations of its mathematical framework, rendering the concept of a superposition of `alive and dead' states more tangible and potentially applicable in practical contexts.

Another foundational principle is Heisenberg's uncertainty principle that asserts a fundamental limit to the simultaneous measurement precision of certain pairs of physical properties, such as position and momentum \cite{Fey_quantum}. Specifically, as the precision of one property increases, the uncertainty of the other proportionally increases.

Additionally, the phenomenon of quantum entanglement represents a critical aspect of quantum mechanics \cite{Bro20}. Entanglement occurs when two or more particles are generated or interact in such a way that their quantum states become interdependent \cite{Hor09}. As a result, the state of one particle cannot be fully described without reference to the state of the other, regardless of the spatial separation between them. This phenomenon challenges classical notions of locality and underscores the unique characteristics of quantum systems.

To further contextualise this discussion, we compare the operation of a traditional digital computer with that of a quantum computer \cite{Nie02}. A classical digital computer relies on bits, which are always in one of two discrete physical states, representing the binary values `0' and `1'. This behaviour is analogous to an on/off light switch. In contrast, a quantum computer employs quantum bits (qubits), which can occupy the states $\ket{0}$ and $\ket{1}$, analogous to the binary states of a classical bit. However, a qubit can also exist in a superposition of these states, represented mathematically as $\ket{\psi} = \alpha\ket{0} + \beta\ket{1}$, where the coefficients $\alpha$ and $\beta$ are complex numbers that satisfy the normalisation condition $|\alpha|^2 + |\beta|^2 = 1$.

From a physical perspective, the state of a qubit can be visualised on the Bloch sphere (Figure~\ref{Fig1}b). When a closed qubit system interacts in a controlled manner with its environment, measurement reveals the probabilities of finding the qubit in either of its basis states. Specifically, for the state $\ket{\psi} = \alpha\ket{0} + \beta\ket{1}$, the measurement probabilities are given by $P_{\ket{0}} = |\alpha|^2$ and $P_{\ket{1}} = |\beta|^2$. This implies that, upon measurement, the qubit collapses to one of its basis states $\ket{0}$ or $\ket{1}$. Graphically, this measurement process corresponds to projecting the qubit state onto one of the coordinate axes of the Bloch sphere (e.g., the $z$-axis in Figure~\ref{Fig1}b).

To demonstrate that light behaves as a wave, scientists illuminate a narrow slit in an opaque screen to observe how light passes through the slit, bends at its edges and spreads out beyond it. This phenomenon is known as diffraction. When the screen contains two narrow slits, the optical waves diffracted by the two slits interact to produce an alternating pattern of light and dark bands, referred to as interference fringes. These experimental results can be reproduced using waves of different natures, including water waves \cite{Fey_quantum}.

However, the same experiment can be performed using electrons (Figure~\ref{Fig1}c). Each electron passing through the slits is registered on a screen as a single bright spot. As more electrons pass through, the individual bright spots begin to cluster, overlap and merge. Ultimately, a double-slit interference pattern emerges, characterised by alternating bright and dark fringes, analogous to the pattern observed in experiments involving optical waves. This result indicates that each individual electron exhibits wave-like behaviour, described by a wave function $\psi$, which passes through both slits simultaneously and interferes with itself before striking the screen.

The square magnitude of the wave function, $|\psi|^2$, represents the probability density of the particle. Accordingly, the alternating peaks and troughs of the wave function of the electron correspond to a quantum probability pattern:~bright fringes indicate a higher probability of finding an electron, while dark fringes indicate a lower probability. Before an electron strikes the screen, its position is not definite but rather probabilistic—it can be found anywhere that the modulus square of the wave function is non-zero. This probability distribution, where multiple states exist simultaneously, is a manifestation of quantum superposition.

\subsection{Menneer-Narayanan Quantum-Theoretic Concept}
In the 1995 technical report `Quantum-inspired Neural Networks' by Tamaryn Menneer and Ajit Narayanan \cite{Men95}, the authors explore the integration of quantum-theoretic concepts into neural network training methodologies. They propose an innovative approach inspired by the many-worlds interpretation of quantum mechanics, aiming to enhance computational efficiency and address problems that traditional neural networks struggle to solve. Although that work initially received relatively little attention, it stands as a pioneering proposal that has significantly influenced subsequent research into quantum neural architectures \cite{Sch14}.

The core idea involves training multiple single-layer neural networks, each on a distinct pattern, rather than training a single network on multiple patterns. The weights from these individual networks are then combined to form a quantum network, where the weights are calculated as a superposition of the individual weights of the networks. This method employs the concept of superposition from quantum theory to potentially improve learning efficiency and problem-solving capabilities.

The authors draw an analogy between their approach and the famous double-slit experiment in quantum mechanics (Figure~\ref{Fig1}b,~c). Just as an electron can exist in a superposition of paths until measured, the quantum-inspired neural network can maintain a superposition of multiple learned patterns until collapsed into a final trained state. This analogy reinforces the idea that quantum-inspired models can explore multiple solutions simultaneously, akin to quantum parallelism \cite{Nie02}.

The authors validate their approach using two microfeature tasks, demonstrating the potential advantages of their quantum-inspired training method. Thus, their work represents an early effort to merge principles from quantum mechanics with neural network training, laying the groundwork for future research in quantum-inspired computational models \cite{Mak24_illusions, Mak24_APL}. In fact, as demonstrated below, at both the theoretical and computational levels, the double-slit experiment and an experiment showcasing the effect of quantum tunnelling through a potential barrier are conceptually similar. This similarity also establishes a connection between the idea of tunneling-based neural networks discussed in this paper and the Menneer-Narayanan quantum-theoretic concept.

\section{Quantum Tunnelling Effect}
\subsection{Theory}
Quantum tunnelling (QT) is a fundamental phenomenon in quantum mechanics that enables particles to pass through potential energy barriers that would be insurmountable under the laws of classical physics \cite{Mes62, Gri04}. This effect arises due to the wave-like nature of quantum particles, described by Schr{\"o}dinger’s equation, which permits nonzero probability amplitudes even in classically forbidden regions, i.e.~areas where a particle does not have sufficient energy to be.

At a microscopic scale, the effect of QT is a direct consequence of the Heisenberg uncertainty principle and the probabilistic interpretation of quantum mechanics. Unlike classical particles, which require sufficient energy to overcome a barrier, quantum particles can `tunnel' through it due to the non-zero probability of their wavefunction extending beyond the boundaries of the barrier. 

Mathematically, this behaviour is captured by the transmission coefficient, which depends on factors such as the barrier width, height and the energy of the particle, and is described by the time-independent Schr{\"o}dinger equation
\begin{equation}
  \label{eq:SE}
  \left[-\frac{\hbar^2}{2m}\frac{d^2}{d x^2} + V(x)\right]\psi(x) = E\psi(x)\,, 
\end{equation}
where $\psi(x)$ is a wave function, $m\approx9.1093837\times10^{-31}$\,kg is the mass of the electron, $\hbar\approx1.054571817\times10^{-34}$\,J$\cdot$s is Plank's constant and $E$ is the energy of the electron. The profile of the potential barrier is 
\begin{equation}\label{eq:Vx}
    V(x) = 
    \begin{cases}
      0 & \text{for }\,x < 0\\
      V_0 & \text{for }\,0 < x \le a\\
      0 & \text{for }\,x > a\,.
    \end{cases}       
\end{equation}

In classical mechanics, a counterpart of this physical system is a marble ball. While a ball with energy $E<V_0$ cannot penetrate the barrier, an electron, behaving as a matter wave, has a non-zero probability of penetrating the barrier and continuing its motion on the other side. Similarly, for $E>V_0$, the electron may be reflected from the barrier with a non-zero probability.

The electron tunnelling behaviour can be quantified by finding the transmission coefficient from the solution of Eq.~(\ref{eq:SE}) for the potential barrier given by Eq.~(\ref{eq:Vx}). The solution of the Schr{\~o}dinger equation can be written as a superposition of left and right moving waves \cite{McQ97}
\begin{equation}
  V(x) =
    \begin{cases}
      \psi_L(x)=A_1e^{\imagi kx}+A_2e^{-\imagi kx}, & x < 0\\
      \psi_C(x)=B_1e^{\imagi\kappa x}+B_2e^{-\imagi\kappa x}, & 0 < x \le a\\
      \psi_R(x)=C_1e^{\imagi kx}+C_2e^{-\imagi kx}, & x > a\,,
    \end{cases}       
\end{equation}
where $\imagi$ is the imaginary unit, $k=\sqrt{2mE/\hbar^2}$ and $\kappa=\sqrt{2m\alpha/\hbar^2}$ with $\alpha=E-V_0$ (the special cases $E=0$ and $E=V_0$ are treated separately). The coefficients $A, B, C$ are found from the boundary conditions at $x=0$ and $x=a$, requiring that $\psi(x)$ and its derivative have to be continuous. Below, omitting the intermediate derivations \cite{McQ97}, we present the expressions for the probability of the electron transmission through the barrier.

For electron energies smaller than the barrier height ($E<V_0$), there is a non-zero transmission probability \cite{McQ97}   
\begin{equation}
  \label{eq:eq1}
  T\vert_{E<V_0} = \left(1-\beta\sinh^2(\kappa_1 a)\right)^{-1}\,, 
\end{equation}
where $\beta=\dfrac{V_0^2}{4E\alpha}$ and $\kappa_1=\sqrt{-2m\alpha/\hbar^2}$. For $E>V_0$
\begin{equation}
  \label{eq:eq2}
  T\vert_{E>V_0} = \left(1+\beta\sin^2(\kappa a)\right)^{-1}\,. 
\end{equation}

Finally, the expression for $E=V_0$ is obtained by taking the limit of $T$ as $E$ approaches $V_0$, resulting in
\begin{equation}
  \label{eq:eq3}
  T\vert_{E=V_0} = \left(1+\frac{ma^2V_0}{2\hbar^2}\right)^{-1}\,. 
\end{equation}

For example, suppose an electron with energy $E = 5$\,eV encounters a barrier of $V_0 = 10$\,eV and width $a = 1$\,nm. Using the expressions above, we can demonstrate that the transmission coefficient will be $T \approx 10^{-10}$, which suggests that tunnelling is highly unlikely but not impossible. If the barrier width were reduced to $0.5$\,nm, the probability would increase significantly, highlighting how nanoscale engineering can control the effect of QT.

The effect of QT has been widely utilised in semiconductor electronic devices \cite{Esa58, Kah67, Cha74, Ion11}, as well as in various spectroscopy \cite{Mod94, Las23} and microscopy techniques \cite{Bin87, Las23}. Furthermore, electron devices exploiting QT have been demonstrated as fundamental building blocks of NC systems \cite{Fen23, Kwo23}. However, in these neuromorphic computers, QT has not been harnessed directly. Instead, the nonlinear dynamics of entire QT-based electron devices and their associated circuits have been employed as a means of computation.

\subsection{Practical Applications of Quantum Tunnelling}
In practice, QT has been exploited in tunnel \cite{Esa58} and resonant-tunnelling \cite{Cha74} diodes. Additionally, certain NC architectures exploited negative differential resistance \cite{Yil13, Ken24}, a hallmark characteristic of tunnel diodes. Notably, systems based on tunnel diodes and other QT-based devices exhibit significantly lower power consumption compared to conventional integrated electronic circuits \cite{Fen23}. Moreover, QT has played a crucial role in the development of scanning tunnelling microscopy (STM) instrumentation \cite{Bin87}. Lastly, QT of individual electrons has been experimentally observed in quantum dots \cite{Yil13}, which serve as essential building blocks of quantum neuromorphic systems \cite{Mar20_2}.

Recent advancements in quantum computing have increasingly harnessed the phenomenon of QT in both quantum annealing and NC architectures \cite{Abe21}. In quantum annealing, tunnelling enables systems to efficiently navigate complex energy landscapes by allowing quantum states to traverse potential energy barriers, thereby facilitating the discovery of optimal solutions in combinatorial optimisation problems. For instance, D-Wave's quantum annealers exploit quantum tunnelling to solve such problems effectively (see Ref.~\cite{Abe21} and references therein).

In the field of NC systems, efforts have been made to emulate the brain's architecture by integrating QT mechanisms. A particular example is the development of neuromorphic Ising machines that utilise Fowler–Nordheim tunnelling annealers \cite{Che24}. These systems employ pairs of asynchronous ON-OFF neurons, with thresholds adaptively adjusted by annealers replicating optimal escape mechanisms, thereby ensuring convergence to ground states in Ising problems. 

Moreover, the effect of QT is essential in quantum biology, influencing enzymatic reactions and energy transfer in photosynthesis \cite{Geo18, Geo19, Geo20}. In this context, it has been demonstrated that QT might play a role in the quantum mind theories discussed above in this text \cite{Geo21, Geo24}.
\begin{figure}[t]
\centering
\includegraphics[width=1.05\columnwidth]{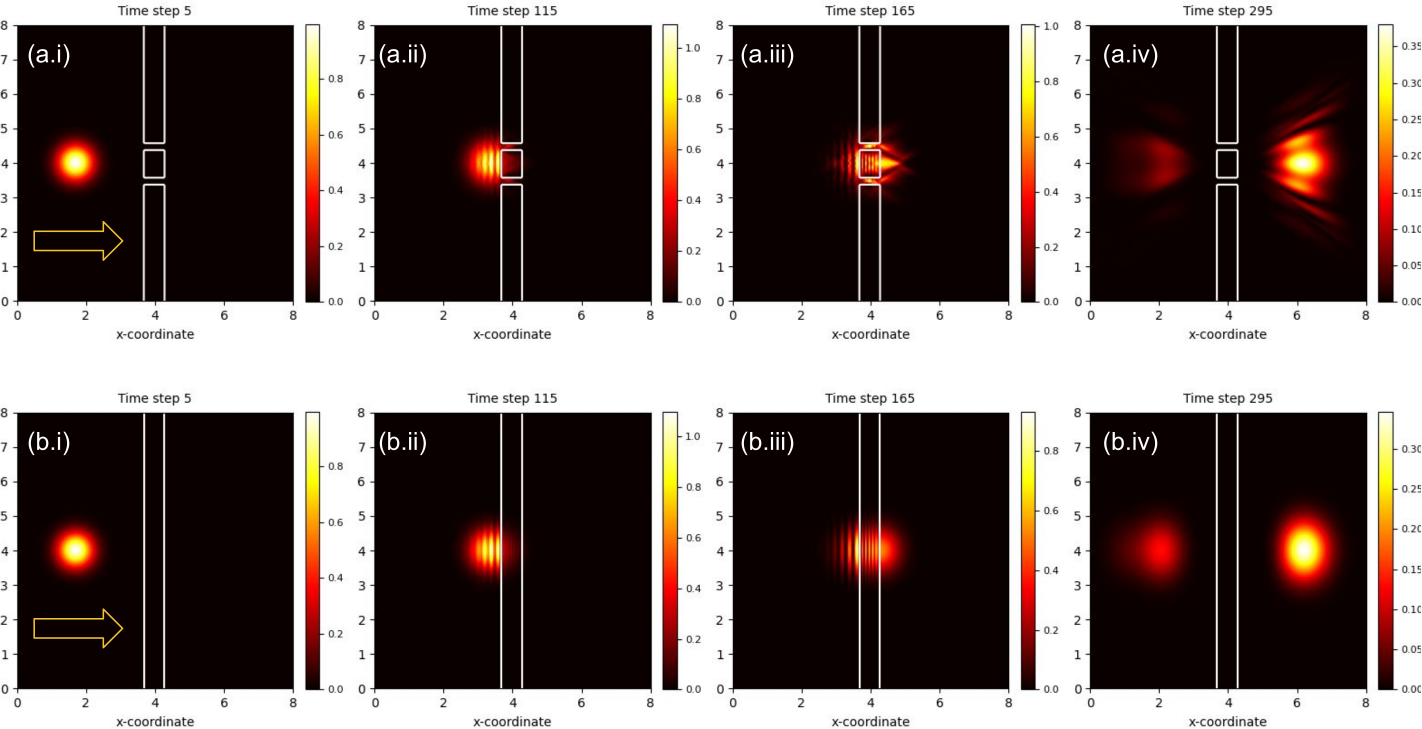}
\caption{{{\bf(a.i)--(a.iv)}~Instantaneous snapshots of an energy wave packet modelling the interaction of an electron with a double-slit structure. \bf(b.i)--(b.iv)}~Instantaneous snapshots of an energy wave packet modelling the tunnelling of an electron through a continuous potential barrier. The false-colour scale of the images encodes the computed probability density values.}
\label{Fig2}
\end{figure}

\subsection{The Relationship between QT and Menneer-Narayanan Quantum-Theoretic Concept}
Let us now focus on Figure~\ref{Fig1}d. In quantum mechanics, the state of an electron remains undefined until its wave function interacts with a detection screen, causing the wave function to `collapse' and the electron to manifest at a specific location. This principle mirrors the approach used by Shor in his quantum computing algorithm for factoring large integers \cite{Sho94}, as well as by Menneer and Narayanan in their pursuit of a robust QNN architecture \cite{Men95}.

As detailed in Ref.~\cite{Men95}, a memory register is initially placed in a superposition of all possible integers it can hold. Following this, a separate calculation occurs in each universe path after the slit. The computation halts when the universes begin to interfere with one another, forming standing waves from the repeating sequences of integers in each universe. While there is no guarantee of accuracy in the results, a subsequent check can confirm if the returned numbers are indeed prime factors of the large integer. Remarkably, quantum computing can solve the prime factorisation problem in seconds, a feat that would take classical computers exponentially longer \cite{Nie02}.

Both the double-slit experiment and QT can be illustrated using a two-dimensional mathematical model, where the electron is represented as a Gaussian-shaped energy packet advancing towards a potential barrier, which may or may not contain slits (Figure~\ref{Fig2}). The motion of the energy packet, with its direction of propagation indicated by the arrows and energy level encoded in false colour, and its interaction with the barrier, represented by white rectangles, are governed by the Schr{\"o}dinger equation, which is solved using the Crank-Nicolson method \cite{Kha22, Mak24_book}.

Figure~\ref{Fig2} plots the probability density in two-dimensional space, illustrating the evolution of the energy packet through four snapshots taken at distinct non-dimensionalised time intervals. The packet then interacts with the barrier, producing both a reflected and a transmitted signal. The top panels (a.i)--(a.iv) of Figure~\ref{Fig2} show the physical picture used by Menneer and Narayanan \cite{Men95}, while the bottom panels (b.i)--(b.iv) offer a more rigorous depiction of the effect of QT compared with the algebraic expressions obtained above, also highlighting the physical similarity between QT-based neural network models and the approach suggested by Menneer and Narayanan.

\section{Benchmarking Neural Network Models}
We begin this section by outlining the traditional neural network algorithms used in this paper as a reference. Then, we present a general framework for transforming these conventional methods into neuromorphic quantum approaches. To demonstrate the effectiveness of this framework, in the following sections we will test the resulting quantum networks on key tasks: classification tasks such as image recognition for FNN, RNN and BNN models, and chaotic time series prediction for the ESN model. For classification tasks involving image recognition, input images will be flattened into vectors and normalised to the range $[0, 1]$, while the corresponding labels will be transformed into one-hot encodings for loss calculation \cite{Kim17}. A similar normalisation of inputs will also be applied in the ESN tests to ensure consistency across tasks \cite{Luk12}.

\subsection{Feedforward Neural Networks}
FNNs are a class of neural networks where information flows in a single direction, from input to output, without feedback connections. They consist of an input layer, one or more hidden layers and an output layer. Each layer is fully connected to the next, with no recurrent connections \cite{Kim17}.

The Rectified Linear Unit (ReLU) activation function is commonly used to introduce nonlinearity to an FNN model by outputting the input value directly if it is positive and producing zero otherwise \cite{Kim17}. However, other activation functions, such as the sigmoid and tanh, can also be employed depending on the specific characteristics of the task, each offering distinct advantages in terms of gradient propagation and model performance \cite{Kim17}.

For a given input vector $x$, the output of the network is computed layer by layer. The activation $a^{(l)}$ of the $l$-th layer is given by the equation
\begin{equation}
a^{(l)} = W^{(l)} z^{(l-1)} + b^{(l)}\,,
\end{equation}
where $W^{(l)} \in \mathbb{R}^{n_l \times n_{l-1}}$ is the weight matrix connecting layer $l-1$ to layer $l$, $z^{(l-1)} \in \mathbb{R}^{n_{l-1}}$ denotes the activations from the previous layer and $b^{(l)} \in \mathbb{R}^{n_l}$ is the bias vector for layer $l$. The output of layer $l$, denoted $z^{(l)}$, is obtained by applying a nonlinear activation function $\sigma$ to the pre-activation $a^{(l)}$ as
\begin{equation}
z^{(l)} = \sigma(a^{(l)})\,.
\end{equation}

At the output layer, the network produces a prediction $\hat{y}$, which depends on the computational task. For classification, the Softmax function is commonly used to compute class probabilities \cite{Kim17}. This function reads
\begin{equation}
\hat{y}_i = \frac{\exp(z^{(L)}_i)}{\sum_{j} \exp(z^{(L)}_j)}\,,
\end{equation}
where $z^{(L)}$ represents the output of the final layer $L$ and $\hat{y}_i$ is the probability of class $i$.

The network is trained by minimising a loss function $\mathcal{L}$, which quantifies the difference between the predicted output $\hat{y}$ and the true target $y$. For classification, the cross-entropy loss is typically used:
\begin{equation}
\mathcal{L} = -\sum_{i} y_i \log(\hat{y}_i)\,.
\end{equation}

To optimise the weights $W^{(l)}$ and biases $b^{(l)}$, gradient backpropagation is employed \cite{Kim17}. This involves computing the gradients of the loss function with respect to the network parameters. Using the chain rule, the gradient $\dfrac{\partial \mathcal{L}}{\partial W^{(l)}}$ is calculated as
\begin{equation}
\frac{\partial \mathcal{L}}{\partial W^{(l)}} = \delta^{(l)} \cdot z^{(l-1)}{\transpose}
\end{equation}
where $\delta^{(l)}$ represents the error at layer $l$, propagated backward through the network
\begin{equation}
\delta^{(l)} = \left(W^{(l+1)}{\transpose} \delta^{(l+1)} \right) \odot \sigma^{\prime}(a^{(l)})\,,
\end{equation}
where $W^{(l+1)}{\transpose}$ is the transpose of the weight matrix from layer $l$ to $l+1$, $\odot$ is the element-wise multiplication operator and $\sigma^{\prime}(a^{(l)})$ is the derivative of the activation function at $a^{(l)}$. Then, the error at the output layer is computed as
\begin{equation}
\delta^{(L)} = \hat{y} - y\,.
\end{equation}

Using these gradients, the parameters are updated via gradient descent
\begin{equation}
W^{(l)} \gets W^{(l)} - \eta \frac{\partial \mathcal{L}}{\partial W^{(l)}}\,,
\end{equation}
\begin{equation}
b^{(l)} \gets b^{(l)} - \eta \frac{\partial \mathcal{L}}{\partial b^{(l)}}\,,
\end{equation}
where $\eta$ is the learning rate parameter.

By iteratively applying these updates over the training data, the network learns to minimise the loss function, improving its performance on the given task.

\subsection{Recurrent Neural Networks}
RNNs are a class of neural networks designed for processing sequential data \cite{Hay01, Jae05, Jun19}. The RNN algorithm relies on the iterative update of a hidden state, thereby capturing information about past inputs in the sequence.

The hidden state at time step $t$, denoted as $h_t$, is updated based on the current input $x_t$ and the previous hidden state $h_{t-1}$ as
\begin{equation}
h_t = \tanh(W_h h_{t-1} + W_x x_t + b_h)\,,
\end{equation}
where $W_h \in \mathbb{R}^{n \times n}$ is the weight matrix for the hidden state, $W_x \in \mathbb{R}^{n \times m}$ is the weight matrix for the input, $b_h \in \mathbb{R}^n$ is the bias vector for the hidden state and $\tanh(\cdot)$ is the hyperbolic tangent activation function introducing nonlinearity to the model.

The output at time step $t$, denoted as $o_t$, is computed based on the current hidden state $h_t$ as
\begin{equation}
o_t = W_y h_t + b_y\,,
\end{equation}
where $W_y \in \mathbb{R}^{k \times n}$ is the weight matrix for the output and $b_y \in \mathbb{R}^k$ is the bias vector for the output.

For classification tasks, the output $o_t$ is processed by means of the Softmax function to compute class probabilities $\hat{y}_t$. In the context of the discussion in this subsection, this function can be written as
\begin{equation}
\hat{y}_t = \text{softmax}(o_t) = \frac{\exp(o_t^{(i)})}{\sum_{j} \exp(o_t^{(j)})}\,,
\end{equation}
where $\hat{y}_t^{(i)}$ represents the probability of the $i$-th class at time $t$.

The loss function over a sequence is computed as the sum of individual losses at each time step is computed as
\begin{equation}
\mathcal{L} = \sum_{t=1}^T \mathcal{L}_t\,.
\end{equation}
The cross-entropy loss that is typically used for classification tasks is
\begin{equation}
\mathcal{L}_t = -\sum_{i} y_t^{(i)} \log(\hat{y}_t^{(i)})\,,
\end{equation}
where $y_t^{(i)}$ is the true label for class $i$ at time $t$.

During the training of the network, gradients of the loss with respect to the weights are computed via Backpropagation Through Time (BPTT) \cite{Hay01}. Gradients for the loss at time $t$ depend on the current hidden state $h_t$ and the previous hidden states $\{h_{t-1}, h_{t-2}, \dots\}$. Due to the repeated application of the chain rule, gradients can either vanish (approach zero) or explode (grow uncontrollably) as they propagate backward through time. This challenge limits the effectiveness of standard RNNs for long sequences and is managed using a gradient clipping technique \cite{Zha20_grad}.

\subsection{Bayesian Neural Networks}
BNNs are a probabilistic extension of traditional neural networks that aim to model the uncertainty in the weights of the network \cite{Jos22, Liu22, Gaw23, Was23}. Instead of learning deterministic weights, BNNs estimate a distribution over the weights. This enables the model to quantify the uncertainty in its predictions, which is particularly useful in situations where decisions must be made under uncertainty. The typical architecture of a BNN consists of an input layer that processes the input data, hidden layers that consist of neurons with probabilistic weights and biases and an output layer that is used to computes the output employing the probabilistic weights.

Let $W_1, W_2$ represent the weights for the first and second layers of the network, respectively. In a Bayesian framework, these weights are not fixed but are treated as random variables with a probability distribution. During training, the model aims to estimate the posterior distribution of the weights given the data.

The weights for the network are parameterised by their mean and standard deviation
\begin{equation}
W_1 = W_{1,\text{mean}} + W_{1,\text{std}} \cdot \epsilon_1, \quad \epsilon_1 \sim \mathcal{N}(0,1)\,,
\end{equation}
\begin{equation}
W_2 = W_{2,\text{mean}} + W_{2,\text{std}} \cdot \epsilon_2, \quad \epsilon_2 \sim \mathcal{N}(0,1)\,,
\end{equation}
where $\epsilon_1$ and $\epsilon_2$ are random variables drawn from a standard normal distribution. The forward pass procedure of the BNN uses the probabilistic weights. The first layer of the network is computed as
\begin{equation}
z_1 = X W_1 + b_1\,,
\end{equation}
where $X$ is the input matrix and $b_1$ is the bias for the first layer. The output of the first layer, $a_1$, is obtained using the ReLU activation function
\begin{equation}
a_1 = \text{ReLU}(z_1)\,.
\end{equation}
The second layer output is then computed as
\begin{equation}
z_2 = a_1 W_2 + b_2\,
\end{equation}
and the output of the network is obtained by applying the Softmax function
\begin{equation}
y_{\text{pred}} = \text{Softmax}(z_2)\,.
\end{equation}

For prediction, the model samples weights from their respective distributions and exploits the sampled weights to make predictions. This process enables the network to estimate the uncertainty in its predictions. The output prediction is the average of predictions made from multiple weight samples
\begin{equation}
y_{\text{pred}} = \frac{1}{N} \sum_{i=1}^N \text{Softmax}(X W_1^{(i)} + b_1^{(i)}, X W_2^{(i)} + b_2^{(i)})\,,
\end{equation}
where $N$ is the number of weight samples and each sample $W_1^{(i)}, W_2^{(i)}$ is drawn from a distribution parametrised by the means and standard deviations of the weights.

The weights of the BNN are trained using a standard stochastic gradient descent algorithm, but with weight uncertainty incorporated into the training process. The gradient of the loss function with respect to the weights is computed using the chain rule and the weights are updated in each step based on these gradients. The loss function is typically computed as
\begin{equation}
\text{Loss} = -\frac{1}{T} \sum_{t=1}^T \sum_{c=1}^C y_t^{(c)} \log(p_t^{(c)})\,,
\end{equation}
where $y_t^{(c)}$ is the true label and $p_t^{(c)}$ is the predicted probability for class $c$ at time step $t$. Then, the gradients with respect to the weights are computed and used to update the means of the weight distributions
\begin{equation}
W_{1,\text{mean}} \leftarrow W_{1,\text{mean}} - \eta \cdot \nabla_{W_1} \text{Loss}\,,
\end{equation}
where $\eta$ is the learning rate and $\nabla_{W_1}$ represents the gradient of the loss with respect to $W_1$. Similar updates are applied for $W_2$, $b_1$ and $b_2$.
\begin{figure}[t]
\centering
\includegraphics[width=0.9\columnwidth]{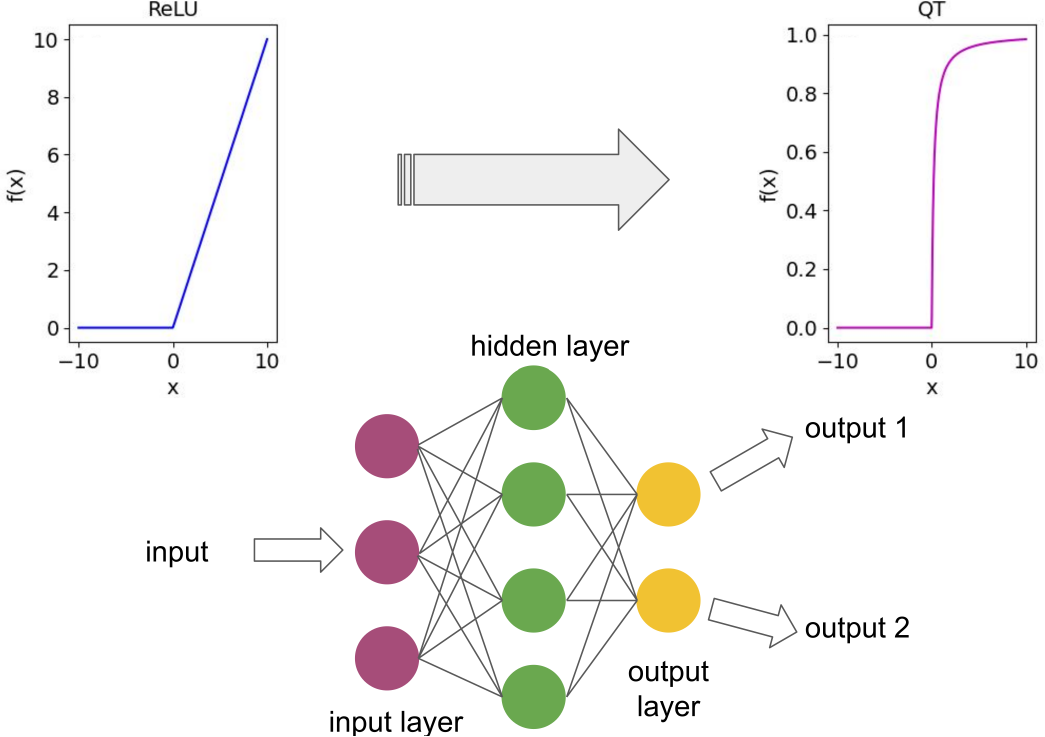}
\caption{Schematic representation of a generic neural network model, illustrating the replacement of the traditional ReLU activation function with the physical QT effect. Other types of activation functions can be substituted in a similar manner. In this paper, the Softmax activation function remains unaltered. The proposed replacement approach has been demonstrated to be effective across all neural network models examined in this study.}
\label{Fig3}
\end{figure}

\subsection{Echo State Networks and Reservoir Computing}
Echo State Networks (ESNs) are an independent class of RNNs specifically designed for processing sequential data, including highly nonlinear and chaotic time-series, while addressing issues such as vanishing and exploding gradients during training \cite{Luk09, Luk12, Tan19, Nak20, Mak23_review}. ESNs exploit a sparsely connected reservoir of dynamic recurrent units with fixed weights, focusing on training only the output weights, which makes it an independent ML technique \cite{Luk09}.
\begin{figure}[t]
\centering
\includegraphics[width=1.0\columnwidth]{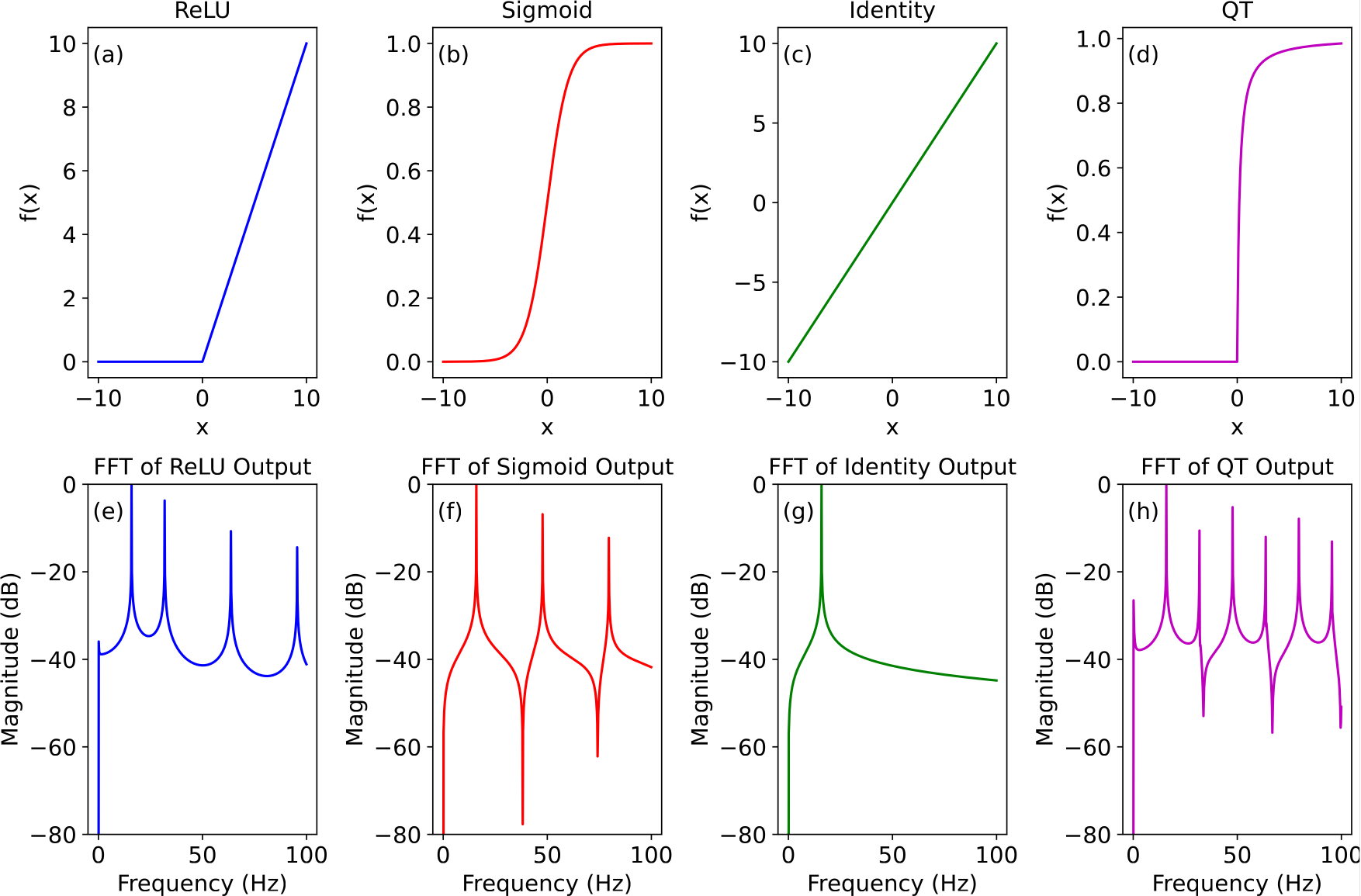}
\caption{Fourier spectra of the outputs of {\bf(a,~e)}~ReLU, {\bf(b,~f)}~Sigmoid, {\bf(c,~g)}~Identity and {\bf(d,~h)}~QT functions activated by a sinusoidal wave signal at a frequency of 16\,Hz. Note the highest number of nonlinearly generated higher-order harmonic in the spectrum of the QT function.}
\label{Fig4}
\end{figure}

A traditional ESN system consists of three primary components:~an input layer that maps the input sequence to the reservoir, a reservoir representing a sparsely connected and randomly initialised network of recurrent neurons, providing rich, in terms of physics and mathematics, nonlinear dynamics \cite{Luk09, Tan19, Mak23_review}, and an output layer that linearly combines the reservoir states to generate predictions.

Similarly to the broader concept of NC, the core concept behind ESN draws inspiration from the functioning of biological brains, which operate through vast, intricate networks of neural connections. Like the brain, neural networks are dynamic systems meaning they evolve over time and exhibit complex, nonlinear and sometimes chaotic behaviour \cite{Mck94, Kor03}. In mathematical terms, dynamical systems are characterised by equations that describe how their states change over time \cite{Mar12}. This similarity between biological and artificial systems has led to the application of principles from nonlinear dynamics in designing ESN-inspired artificial neural network models \cite{Maa02, Jae04}. In particular, nonlinear differential equations are often used to model how the connection strengths between nodes in artificial neural networks evolve over time \cite{Maa02, Jae04, Luk09}.
\begin{figure}[t]
\centering
\includegraphics[width=1.0\columnwidth]{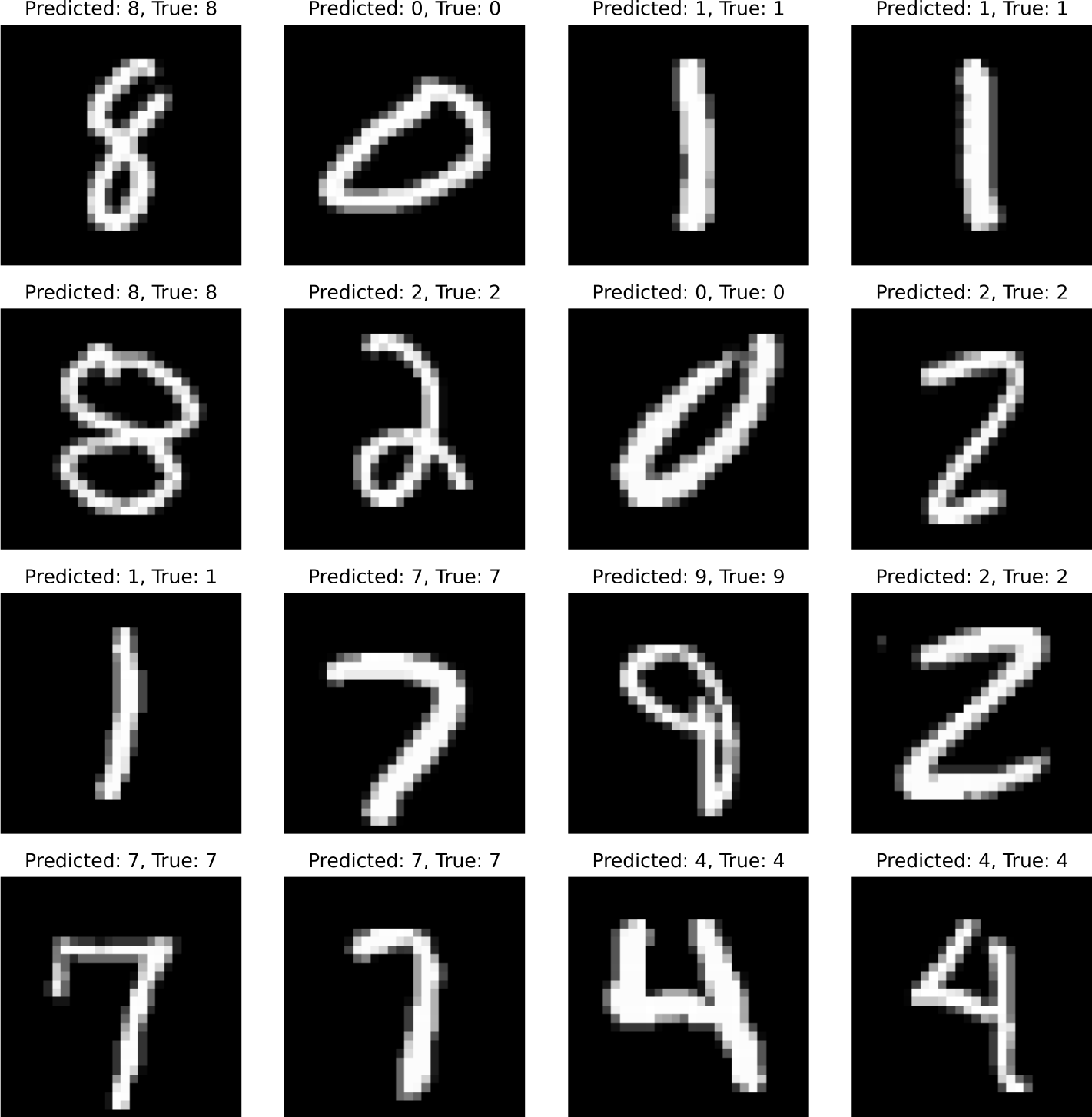}
\caption{Example of classifications from a randomly selected subset of the MNIST testing dataset produced by the QT-feedforward neural network. The labels above each panel indicate the predicted categories alongside the ground truth labels.}
\label{Fig5}
\end{figure}

For a given input temporal sequence $\{x_t\}_{t=1}^T$, the dynamical state of the reservoir $h_t$ at time step $t$ is updated as
\begin{equation}
h_t = \tanh(W_{\text{in}} x_t + W_{\text{res}} h_{t-1} + b)\,,
\end{equation}
where $W_{\text{in}} \in \mathbb{R}^{N \times M}$ is the input weight matrix, $W_{\text{res}} \in \mathbb{R}^{N \times N}$ is the reservoir weight matrix, $b \in \mathbb{R}^N$ is the bias vector, $N$ is the number of reservoir neurons and $\tanh(\cdot)$ is the hyperbolic tangent activation function.

The reservoir weights $W_{\text{res}}$ are scaled to ensure the echo state property, which guarantees that the states of the network are stable and dependent on the input history \cite{Luk09, Luk12}. This is typically achieved by setting the spectral radius $\rho$ of $W_{\text{res}}$ to be less than 1:
\begin{equation}
\rho = \max |\lambda|\,,
\end{equation}
where $\lambda$ are the eigenvalues of $W_{\text{res}}$. The output at each time step, $y_t$, is computed as
\begin{equation}
y_t = W_{\text{out}} h_t\,,
\end{equation}
where $W_{\text{out}} \in \mathbb{R}^{K \times N}$ is the output weight matrix and $K$ is the dimension of the output. Unlike traditional RNNs, ESNs train only the output weights $W_{\text{out}}$, keeping $W_{\text{in}}$ and $W_{\text{res}}$ fixed. The training involves solving a linear regression problem
\begin{equation}
W_{\text{out}} = Y H^\dagger\,,
\end{equation}
where $Y \in \mathbb{R}^{K \times T}$ is the desired output matrix, $H \in \mathbb{R}^{N \times T}$ is the matrix of reservoir states over time and $H^\dagger$ is the pseudo-inverse of $H$ \cite{Luk12}. Regularisation, such as ridge regression \cite{Luk12}, is often used to prevent overfitting
\begin{equation}
W_{\text{out}} = Y H^\top (H H^\top + \lambda I)^{-1}\,,
\end{equation}
where $\lambda$ is the regularisation coefficient.

\subsection{From Classical to Quantum: Transforming Computational Models}
The process of converting a traditional neural network model into a QT-based model is relatively straightforward: the conventional activation functions (as exemplified by ReLU in Figure~\ref{Fig3}) are replaced by the physical QT effect. In this work, the Softmax function remains unchanged, although some success has been achieved in experimental tasks involving fully QT-based neural networks.

The same procedure applies to the derivative of the activation function, if it is used in the particular model of interest \cite{Mak24_APL}. Additionally, applying the QT activation function and its derivative to ML algorithms may require basic mathematical normalisation to regulate the numerical values entering the algorithm (see Ref.~\cite{McN25} for a relevant discussion). This procedure is heuristic and programmatically straightforward (and the variable {\tt ampl} used in the source code accompanying this paper serves this purpose).

Naturally, the replacement procedure illustrated in Figure~\ref{Fig3} alters the dynamics of neural network training and deployment, necessitating a readjustment of the mathematical range of the activation function, weight distribution, number of neurons in the hidden layer, learning rate and the number of epochs and batches. However, we have established that satisfactory performance can be achieved simply by replacing the traditional activation functions with QT, provided that the original traditional neural network model was appropriately tuned for the specific task at hand. Importantly, we also demonstrated that the resulting QT-based model can be trained 50 times faster without any additional adjustments and holds potential for further acceleration with proper tuning \cite{Maks25}.

Aside from their empirical performance, activation functions also possess distinct mathematical properties, including the necessity of nonlinearity as stipulated by the universal approximation theorem \cite{Cyb89, Hay98, Kim17}. In this paper, we demonstrate that the QT activation function exhibits a greater degree of nonlinearity compared to the standard ReLU and sigmoid activation functions (Fig.~\ref{Fig4}).
\begin{figure}[t]
\centering
\includegraphics[width=0.7\columnwidth]{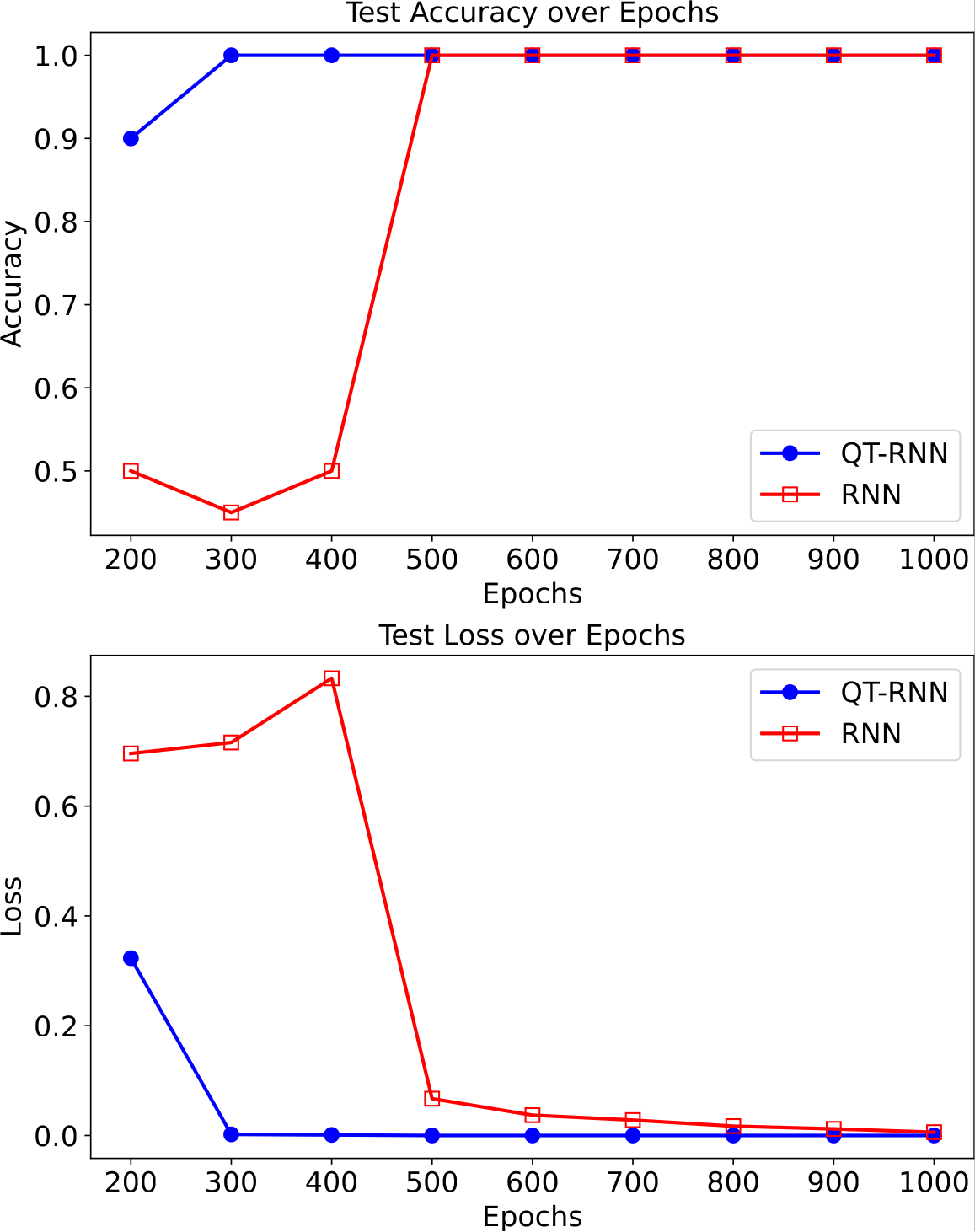}
\caption{Test accuracy and loss over training epochs for QT-RNN (blue circles) and standard RNN models (red squares).}
\label{Fig6}
\end{figure}

Analysing the degree of nonlinearity of an activation function is a non-trivial task. A previous study \cite{Mak19} suggested that two nonlinear processes can be compared by examining the Fourier spectra of the functions that approximate these processes. Following this approach, we analyse the responses of the ReLU, sigmoid and QT activation functions to a purely sinusoidal wave signal at a frequency of 16\,Hz. For the sake of comparison, we also apply the sinusoidal signal to a linear identity function, producing a Fourier spectrum with a sole peak at 16\,Hz. 

As shown in Fig.~\ref{Fig4}, the responses of ReLU, sigmoid and QT to the sinusoidal waves result in the nonlinear generation of higher-order harmonics. According to \cite{Mak19}, the strength of the nonlinearity can be quantified using the magnitude of the harmonic peaks and the total number of harmonics produced by the nonlinear process. Focusing on the latter criterion, we can see that ReLU produces the peak at the second-harmonic frequency 32\,Hz and then at the fourth, 64\,Hz, and the sixth, 96\,Hz, harmonics. In turn, sigmoid produces the third, 48\,Hz, and fifth, 80\,Hz, peaks only. On the contrary, QT generates strong peaks at both odd and even harmonic frequencies. Within the theoretical framework used in this analysis, this result suggests that the QT exhibits the strongest nonlinearity among the three analysed functions. We also note that the nonlinearity of QT can be enhanced by using higher, yet physically realistic, values of its model parameters such as the thickness of the potential barrier.
\begin{figure}[t]
\centering
\includegraphics[width=1.0\columnwidth]{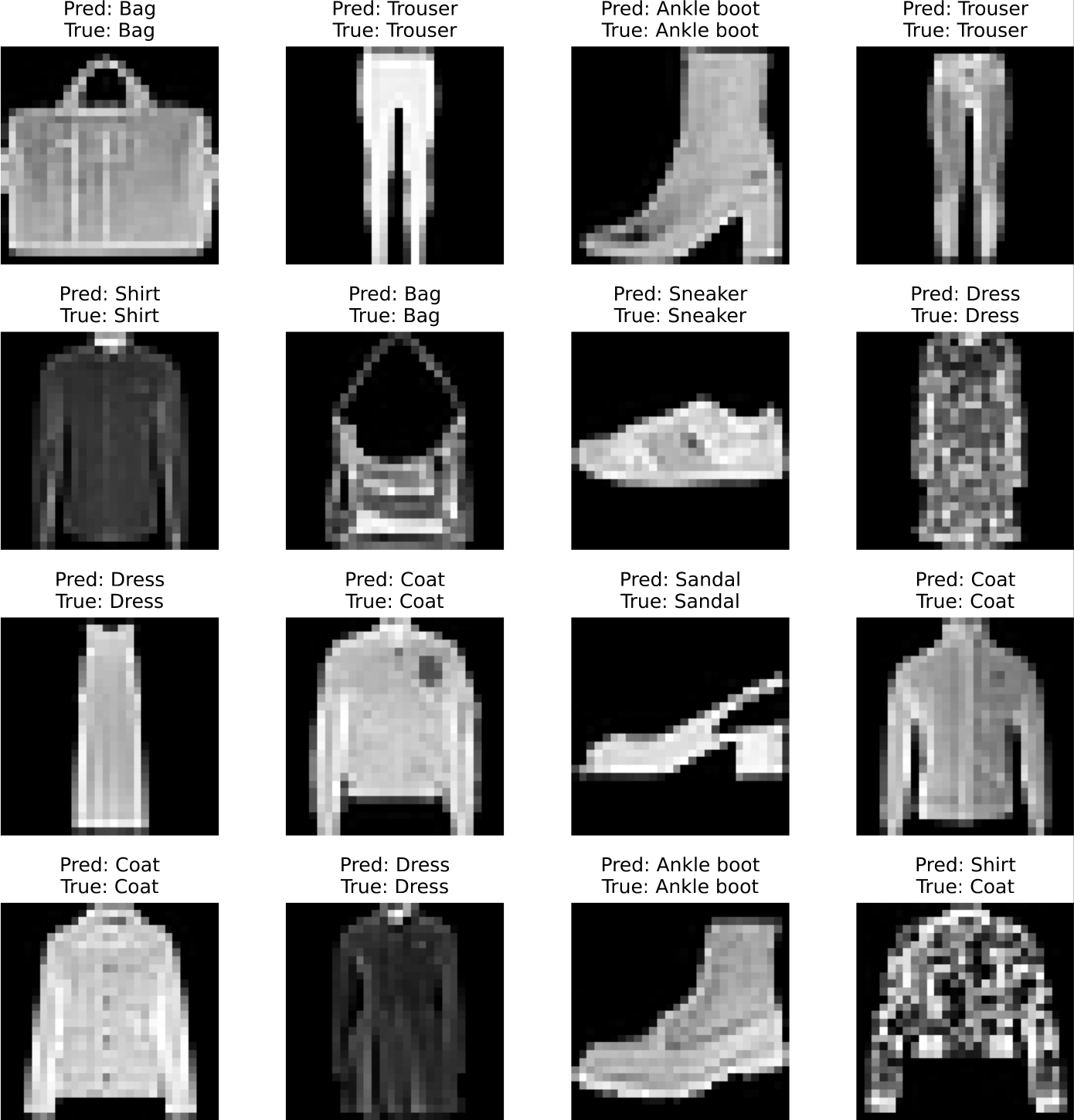}
\caption{Example of classifications from a randomly selected subset of the Fashion MNIST testing dataset produced by the QT-BNN model. The labels above each panel indicate the predicted categories alongside the ground truth labels.}
\label{Fig7}
\end{figure}

\section{Results and Discussion}
\subsection{QT-Feedforward Neural Network}
In this section, we task the QT-based model to classify images from the MNIST (Modified National Institute of Standards and Technology) dataset. This dataset is a widely used benchmark in computer vision and ML research due to its balanced class distribution and moderate computational requirements. It contains 60~000 greyscale images of size 28$\times$28 pixels, categorized into 10 classes representing handwritten digits from 0 to 9, with 6000 images per class.

The model used in this study comprises a single hidden layer consisting of 512 neurons. The learning rate is set at 0.01, with training conducted over 100 epochs. The batch size is configured to 64 and a gradient clipping value of 5 is applied to mitigate the effects of exploding gradients during the training process.

In Figure~\ref{Fig5}, the model demonstrates accurate classification of all test images. The overall accuracy achieved during training in this configuration is 98.7\%, while the testing of all images from the test section of the MNIST dataset yielded an accuracy of 98.3\%.

It is worth noting that fully connected neural networks are generally suboptimal for MNIST image classification tasks because they do not explicitly explore the spatial correlations present in image data. Convolutional neural networks (CNNs), on the other hand, are specifically designed to capture local patterns and hierarchical features through convolutional layers, making them more effective for such tasks. However, the QT-based network, despite its fully connected architecture, demonstrates decent performance on MNIST images. The incorporation of a QT activation function introduces a novel nonlinearity that helps mitigate some of the inherent limitations of standard fully connected layers. Although the QT network may not reach the optimal performance levels of state-of-the-art CNNs and other advanced models, its ability to deliver respectable accuracy suggests that physics-inspired modifications can offer a viable alternative approach for enhancing fully connected models in complex image classification scenarios.

\subsection{QT-RNN for a Sentiment Analysis Task}
In this section, we demonstrate the transformation of a traditionally built RNN sentiment analysis model \cite{Zho19} into a QT-based model. The original model processes a dataset structured as a two-column table, where the first column contains text phrases and the second column classifies them as positive or negative \cite{Zho19}. The dataset is divided into two subsets: a training set for model learning and a testing set for performance evaluation.

First, we construct a vocabulary encompassing all unique words in the dataset and assign each word a corresponding integer index. The model is then trained on the training subset and evaluated on the testing subset to assess its classification accuracy.

Figure~\ref{Fig6} presents the test accuracy and loss over training epochs for QT-RNN and standard RNN models. We can see that the QT-RNN (blue circles) achieves rapid convergence, reaching 100\% accuracy with near-zero loss after 300 epochs. In contrast, the traditional RNN (red squares) shows slower improvement, initially struggling with accuracy fluctuations and higher loss before eventually converging. These results highlight the advantages of the QT-based approach in optimising model performance \cite{Maks25}.

In addition to faster convergence, the near-zero loss values from early epochs indicate that QT-RNN should be more stable during training. This stability can be a result of architectural choices (such as gating mechanisms or normalisation techniques \cite{Mak24_APL, Maks25}) that help control gradient issues often seen in traditional RNNs. Arguably, QT-RNN could also be employing an architecture that captures the underlying patterns in the data more efficiently. This property can result from enhanced nonlinearity (Figure~\ref{Fig4}) or mechanisms that better preserve long-term dependencies, leading to more robust feature learning \cite{Maks25}. Moreover, the superior performance of QT-RNN may indicate that the quantum model possesses an increased ability to generalise, which results in better performance on test data.

\subsection{QT-Bayesian Neural Network}
To test the performance of the QT-BNN model, we employ the Fashion MNIST dataset that consists of 70~000 greyscale images (28$\times$28 pixels) representing 10 categories of fashion items, such as T-shirts, trousers and sneakers. Fashion MNIST was created as a more challenging alternative to the classic MNIST dataset of handwritten digits. While MNIST has been a {\it de facto} standard for benchmarking ML models, it is often considered too simplistic for modern algorithms, which can achieve near-perfect accuracy. Fashion MNIST, on the other hand, introduces greater complexity due to the variability in textures, shapes and patterns within each class, making it harder for ML models to distinguish between similar categories like shirts, coats and pullovers. This increased difficulty makes Fashion MNIST a more realistic and practicable benchmark for evaluating the performance of ML models, particularly in tasks like image classification and object recognition, where real-world data is often noisy and ambiguous.

The model used in this study is a relatively simple yet efficient QT-based BNN architecture designed for classification tasks on the Fashion MNIST dataset. It consists of a single hidden layer with 512 neurons, which provides a balance between model capacity and computational efficiency. The learning rate is set to 0.5, a relatively high value that allows for faster convergence during training. Training is conducted over 1000 epochs. To simplify the software implementation, no batching was used during training, meaning the model updates its weights after processing each individual sample. For the Bayesian component of the algorithm, 50 samples were used to approximate the posterior distribution. While this is a modest number for a Bayesian model, it strikes a balance between computational efficiency and the ability to capture uncertainty in the predictions. 

Overall, this experimental setup provides a clear yet rigorous framework for evaluating the performance of BNNs on the Fashion MNIST dataset, as demonstrated by the results in Figure~\ref{Fig7}. In this instance, the model exhibited strong classification performance, correctly identifying all items except for a single misclassification, where a coat was predicted as a shirt. This type of error is scientifically relevant, as it highlights the inherent challenges associated with distinguishing visually similar classes within the dataset \cite{Maks25}. Indeed, coats and shirts, for example, may share common visual features such as texture, shape or pattern, particularly in greyscale images where colour information is absent. Therefore, the ability of the QT-BNN model to adequately process an image dataset with such characteristics makes this model valuable for undertaking real-world classification tasks \cite{Maks25}.
\begin{figure}[t]
\centering
\includegraphics[width=1.0\columnwidth]{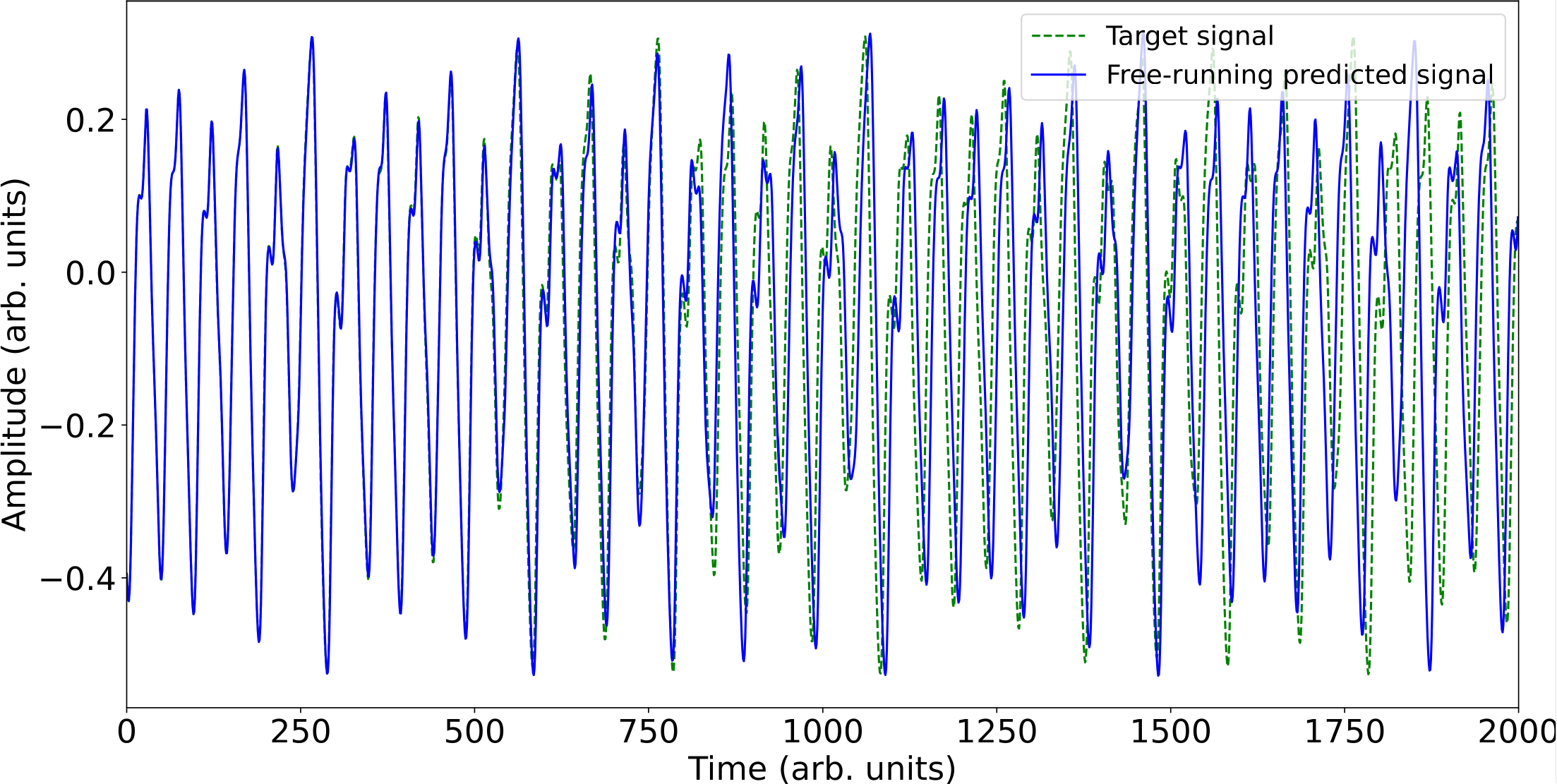}
\caption{Forecast of MGTS made by the QT-ESN system (blue solid line) compared with the target signal (green dashed line).}
\label{Fig8}
\end{figure}

\subsection{QT-ESN and Mackey-Glass Time Series Forecast Task}
In this section, we use a well-documented and highly optimised source code that accompanies the work Ref.~\cite{Luk12} and convert it into a QT-based model. The RC system contains 1000 neurons and in it is trained to forecast the future evolution of a Mackey-Glass time series (MGTS), a standard and relatively challenging task for ML systems \cite{Luk12, Mak23_review, Abb24}.

We generate an MGTS dataset solving the delay differential equation \cite{Mac77}
\begin{eqnarray}
  \dot{x}_{_{MG}}(t)
  &=&\beta_{_{MG}}\frac{x_{_{MG}}(t-\tau_{_{MG}})}
      {1+x_{_{MG}}^{q}(t-\tau_{_{MG}})}-\gamma_{_{MG}}x_{_{MG}}(t)\,,
  \label{eq:MG}
\end{eqnarray}
where overdot denotes differentiation with respect to time and $\tau_{_{MG}}=17$, $q=10$, $\beta_{_{MG}}=0.2$ and $\gamma_{_{MG}}=0.1$ \cite{Luk12}. The parameters used in this equation ensure that the time series exhibits highly nonlinear and also chaotic behaviour. We split the so-generated dataset into two equal parts, each 2000 discrete time steps long. The first part is used to train the RC system but the second one is used as the target data that are not known to the RC system but used as the ground truth exclusively to evaluate the accuracy of the forecast made by the RC system.

Known as the free-running mode of operation, this regime transforms the RC system into a generator of the future temporal evolution of MGTS \cite{Mak23_review}. Naturally, the predicted signal will deviate from the target (ground truth) after a certain period of time. The accuracy of the forecast is measured as the mean squared error between the predicted outputs and the target values as
\begin{equation}
 \text{NMSE} = \frac{1}{N} \sum_{i=1}^{N} \left(y_i^{\text{target}} - \hat{y}_i \right)^2\,,
\end{equation}
where $N$ is the number of testing samples.

Figure~\ref{Fig8} presents the forecast generated by the QT-ESN system (blue solid line) alongside the ground truth target signal (green dashed line). The QT-ESN output remains virtually indistinguishable from the target signal within the time range from 0 to 500. The NMSE, computed over the entire time range from 0 to 2000, is 4.2$\times$10$^{-5}$---only an order of magnitude higher from the highly optimised traditional ESN. Given that the QT-ESN was constructed solely by replacing the traditional tanh activation function with the QT function, without any further optimisation, this result provides strong evidence for the plausibility of the QT activation function as a high-performance activation function for RC systems.    
\begin{figure}[t]
\centering
\includegraphics[width=1.0\columnwidth]{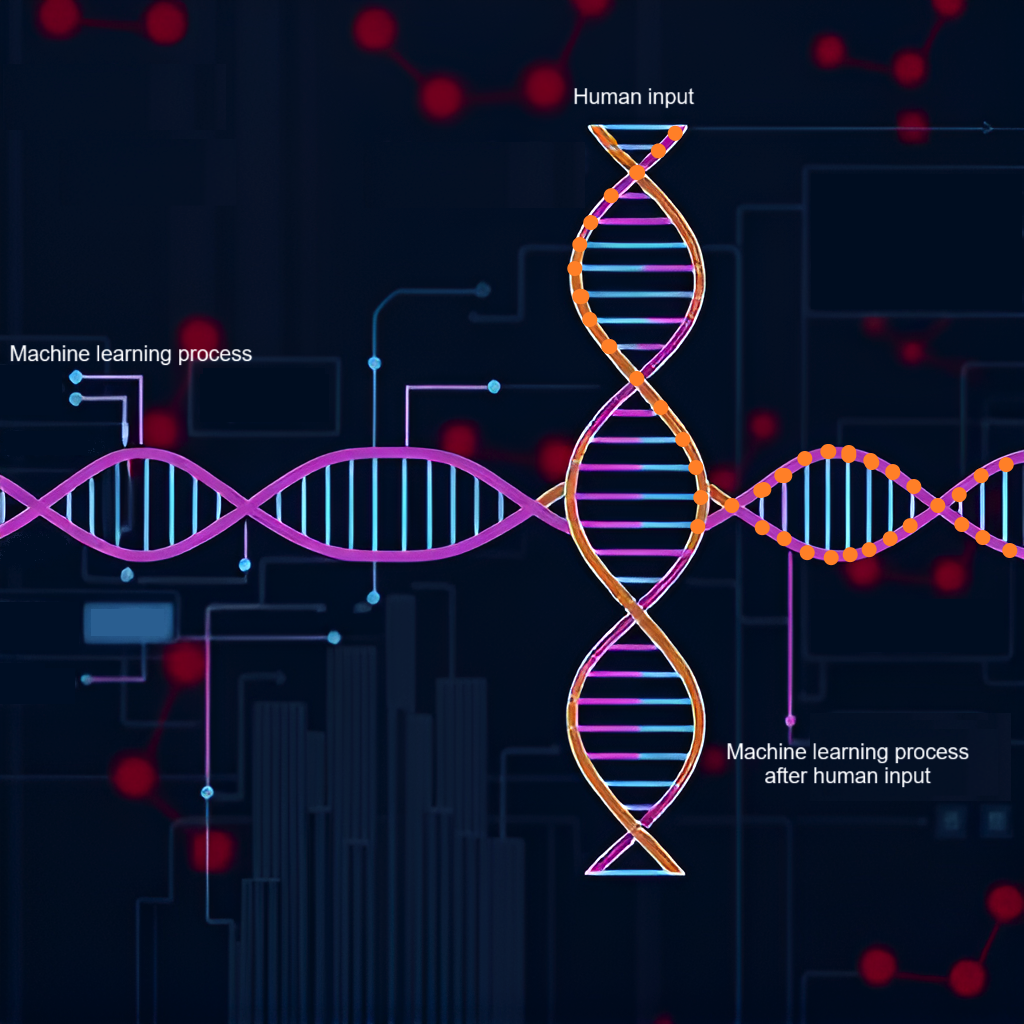}
\caption{Conceptual illustration of Cognitive Human-Machine Teaming, highlighting cognitive interactions between human operators and ML processes. The bright dots emerging from the human input represent the flow of information. Own work by the first author.}
\label{Fig9}
\end{figure}

\section{Potential Applications of QT-Based Models in Neuromorphic and Cognitive AI}
Thus far, we have demonstrated the successful application of QT-based neural network models to standard tasks in machine vision, language processing and chaotic time-series forecasting. These results establish the viability of QT-based architectures as robust and efficient alternatives to conventional neural networks in existing AI applications. However, the potential of QT-based networks extends far beyond these benchmarks, especially due to a direct relevance of the effect of QT to the framework of the quantum cognition and decision-making theory \cite{Bus12, Mak24_illusions, Mak24_APL}. In particular, in what follows we will aim to conceptually demonstrate that their ability to process information through quantum-inspired mechanisms, including those illustrated in Figure~\ref{Fig1}, could influence the field of human-machine teaming.

Human-machine teaming, where humans and machines collaborate to enhance each other's strengths (Figure~\ref{Fig9}), is an evolving field with significant applications in military operations, medicine and AI development \cite{McN18, Lyo19, Hen22, Gre23}. QT-based neural networks, which more closely align with human cognitive processes \cite{Mak24_APL, Maks25}, have the potential to make advances in this domain. By enabling AI systems to interpret complex data while exhibiting adaptive reasoning, situational awareness and real-time strategic planning, these networks can go beyond conventional machine intelligence \cite{Mak24_APL, Maks25}. Yet, we argue that their ability to navigate uncertainty and make rapid yet reliable decisions should make them particularly valuable for defence and security applications, where high-stakes environments demand both precision and agility (the relevant results will be published elsewhere).  

\subsection{Cognitive Human-Machine Teaming}
Good practical application design reduces the cognitive load on operators and helps improve their situational awareness \cite{Tzafestas2010, Wu2016, Yan2017}. This is particularly important in environments where quick and accurate decision-making is required, such as in defence operations \cite{LaurilaPant2023, Lim2018, Pei2024, Zhu2023}. AI systems can support critical operations by reducing cognitive load and improving decision-making speed, which in turn can lead to higher mission success rates \cite{Brachten2020, Carvalho2020}. For example, in our previous work, we demonstrated that algorithms similar to those considered in this paper can make real-time decisions about commands received by an autonomous vehicle from both human operator and physical processes occurring in the surrounding environment \cite{Abb24_1}. Such a functionality has the potential to significantly reduce the cognitive load on the drone operator, thereby decreasing the likelihood of errors. Because the commands given by a human to the drone can be represented as a time-dependent sequence of voltage pulses, the same AI system can be configured to learn from the past inputs of the operator to generate new inputs in a manner consistent with the operator's style of controlling the drone.

Identifying cognitive demands means looking closely at the tasks people perform, the decisions they must make and the overall cognitive workload they experience in different scenarios \cite{McDermott2017, Carvalho2020}. When designing systems for complex environments, whether on the battlefield or in high-tech settings like AI algorithms, it is important to understand the cognitive efforts users must experience. By breaking down the tasks and decisions involved, designers can create interfaces and tools that help manage cognitive load rather than adding to it, a process that involves examining operators’ work and understanding how each task contributes to their overall cognitive burden \cite{McDermott2017, Madni2018}.

A practical way to address these issues is through cognitive task analysis, which maps out the exact steps and decisions required in an operation. For example, in military and spaceship control settings, using cognitive task analysis has provided valuable insights that lead to systems better tuned to human needs \cite{McDermott2017}. Similarly, by assessing cognitive workload through established methodologies, designers can adjust system requirements to more closely align with what human operators need to function efficiently \cite{Madni2018}. In fact, as discussed in more detail in Ref.~\cite{Mak24_APL}, QT-based neural network models can help establish to which extent a human operator has been affected by adverse environments such as weightlessness. 

As part of practical application, experiments can be designed to observe how different versions of neural network algorithms affect performance measures such as prediction accuracy and classification outcomes, focusing on how closely they mimic human decision-making through quantum cognition and decision theories \cite{Yearsley2016, Maks25}. In line with this approach, this paper presents a ready-to-use framework for benchmarking traditional and quantum neural network models that account for the effects of QT, enabling the investigation of AI-driven decision-making in defence applications and beyond. Specifically, this toolbox provides an opportunity to examine the interplay between quantum mechanics and its application in AI by evaluating the impact of varying QT parameters (e.g. the width and height of the potential barrier---each directly relevant to the models of cognitive processes \cite{Mak24_information}) on model predictions and classification outcomes, with a focus on emulating human-like reasoning \cite{Maks25}.

We suggest that experiments should be conducted to run in parallel multiple QT algorithms with different configurations, which is a computationally affordable task given the relative technical simplicity of the codes that accompany this paper. These algorithms should be able to pass information to each other and learn from each other, mimicking human conscience and subconscious interactions, which is also a computationally affordable task. Thus, quantum ML would benefit from being able to run even more processes at different levels than humans could ever be able to do (Figure~\ref{Fig9}).

Practical applications of the outlined approach could include building an experiment for military object detection, such as identifying military tracks, where military drones might use multiple cameras installed at different angles (e.g., 30\degree~left, right, top and bottom) in addition to central camera video capture. Work on such datasets is currently in progress and the scientific findings will be published separately.

Multimodal models, which process and integrate multiple data types simultaneously, offer significant advantages by combining inputs such as audio, text and video to generate more accurate and insightful results \cite{Lia24}. This article presents examples of how the QT-based model can be applied to images, text and time-dependent data, with the latter encompassing not only complex time series but also potentially speech and music patterns \cite{MdR23}. Therefore, these algorithms can be further expanded to process audio and video inputs, not only from human operators but also from cognitive interactions between humans and machines \cite{Haigh2023}.

An advanced system should also integrate cognitive interactions from both humans and machines, fostering effective human-machine teaming through communication via text and voice. In this context, practical applications must be designed to learn from environmental cues provided by both the system and its users, creating a dynamic feedback loop where the cognitive system continuously adapts to human input while users refine their interactions with the system \cite{Haigh2023}. A schematic illustration of such an interaction is presented in Figure~\ref{Fig9}, potentially encompassing the multimodal capabilities of the QT-based neural network models demonstrated in this paper.

Such initiatives in cognitive human-machine teaming underscore the increasing demand for innovative computing paradigms, often termed cognitive computers \cite{Wan24}. In response, quantum and NC systems are establishing the groundwork for the next generation of AI and ML models, enabling more adaptive, efficient and human-like decision-making in complex environments \cite{Mar20_2, Maks25}. Consequently, we anticipate that the paradigm of QT-based neural network models will find a distinct niche within this rapidly evolving field.

\section{Conclusions}
Quantum technologies are rapidly reshaping multiple scientific and industrial domains, playing an increasingly prominent role in fields such as computing, secure communication, sensing and even medical diagnostics. While quantum computing promises revolutionary advancements, its accessibility remains a significant challenge, primarily confined to research institutions and specialised industries with access to high-performance quantum-physical hardware. This paper addresses this gap by demonstrating how traditional neural networks can be transformed into neuromorphic quantum models, significantly lowering the barrier to entry for quantum-inspired computing.

By applying the fundamental principles of quantum mechanics, we have shown that widely used neural network architectures---including feedforward neural networks, recurrent neural networks, reservoir computing models such as the echo state network and Bayesian neural networks---can be adapted to improve both computational efficiency and cognitive-like processing capabilities. The quantum-inspired versions of these models exhibit key advantages, such as more efficient training and an ability to capture aspects of human-like reasoning. This approach not only improves the performance of AI models but also provides insights into the intersection of quantum mechanics and cognitive computing.

One of the most significant contributions of this work is its emphasis on accessibility. Unlike conventional quantum computing frameworks that require specialised hardware and deep technical expertise, the quantum-inspired models presented in this paper can be implemented using standard computational resources, such as a laptop, with only an undergraduate-level understanding of ML. Making these methods available to a wider audience opens new possibilities for researchers, engineers and students alike, enabling a broader community to explore and apply quantum principles in AI.

Beyond conventional ML tasks, our findings suggest that quantum-inspired networks have the potential to transform fields such as human-machine collaboration, adaptive decision-making and cognitive AI. By embedding quantum dynamics into neural architectures, we move closer to developing AI systems that not only process information efficiently but also demonstrate traits of flexible reasoning, contextual awareness and adaptive learning—key elements for the next generation of intelligent systems.

Future research should further investigate the scalability of these models in real-world applications, including defence, security, healthcare and autonomous systems. Additionally, exploring how these networks can be integrated with emerging quantum hardware could lead to hybrid classical-quantum AI frameworks with unprecedented computational power. Ultimately, this work marks a step toward making quantum-inspired neural networks more practical, accessible and impactful, bringing us closer to realising the full potential of quantum technologies in everyday AI applications.

\vspace{6pt} 

\authorcontributions{ISM developed the traditional and quantum neural network models used in this study and obtained the primary results. The concept for the QT-BNN algorithm originated from MM, who also envisioned the application of QT-based neural network models in cognitive human-machine teaming and authored the relevant sections of the paper. ISM edited the manuscript as a whole, with contributions from MM.}

\funding{This research received no external funding}

\institutionalreview{Not applicable.}

\informedconsent{Not applicable.}

\dataavailability{This paper has no additional data. The source codes that accompany this paper are available in the GitHub repository, \url{https://github.com/IvanMaksymov/Quantum-Tunnelling-Neural-Networks-Tutorial}.} 


\conflictsofinterest{The authors declare no conflicts of interest.} 



\abbreviations{Abbreviations}{
The following abbreviations are used in this manuscript:\\

\noindent 
\begin{tabular}{@{}ll}
AI & artificial intelligence\\
BNN & Bayesian neural network\\
BPTT & Backpropagation Through Time\\
CNN &  convolutional neural networks\\
ESN & Echo State Network\\
FNN & feedforward neural network\\
MGTS & Mackey-Glass time series\\
ML & machine learning\\
MNIST & Modified National Institute of Standards and Technology database\\
NC & neuromorphic computing\\
QT-BNN & quantum-tunnelling Bayesian neural network\\
QT-ESN & quantum-tunnelling Echo State Network\\
QT-FNN & quantum-tunnelling feedforward neural network\\
QT-RNN & quantum-tunnelling recurrent neural network\\
QCT & quantum cognition theory\\
QNN & quantum neural network\\
QT & quantum tunnelling\\
RNN & recurrent neural network\\
ReLU & rectified linear unit\\
STM & scanning tunnelling microscopy\\
\end{tabular}
}

\begin{adjustwidth}{-\extralength}{0cm}

\reftitle{References}


\bibliography{refs}

\end{adjustwidth}
\end{document}